\pdfoutput=1

\documentclass[11pt]{article}

\usepackage[preprint]{acl}

\usepackage{times}
\usepackage{latexsym}

\usepackage[T1]{fontenc}

\usepackage[utf8]{inputenc}

\usepackage{microtype}

\usepackage{inconsolata}

\usepackage{graphicx}

\usepackage{graphicx}
\usepackage{subfig}
\usepackage{amssymb}
\usepackage[acronym]{glossaries}
\glsdisablehyper
\makeglossaries

\newacronym{llm}{LLM}{large language model}
\newacronym{mc}{MC}{multiple choice}
\newacronym{qa}{QA}{question answering}
\newacronym{ll}{LL}{log-likelihood}
\newacronym{nlu}{NLU}{Natural Language Understanding}
\newacronym{nlg}{NLG}{Natural Language Generation}
\newacronym{CoT}{CoT}{chain-of-thought}

\usepackage{multirow}
\usepackage{pgfplots}
\pgfplotsset{compat=1.17}

\usepackage{todonotes}

\usepackage{xcolor}
\usepackage{ifthen}
\usepackage{longtable}
\usepackage{verbatim}
\usepackage{enumitem}
\usepackage{makecell}
\usepackage{tikz}
\usepackage{array}
\newif\ifshowtodos
\showtodostrue   
\newif\ifshowdg
\showdgtrue 
\newif\ifshowvh
\showvhtrue

\newcommand{\valuewithsd}[2]{#1\textcolor{gray}{\scriptsize(#2)}}
\newcommand{\tablevaluewithsdsix}[6]{\valuewithsd{#1}{#2} & \valuewithsd{#3}{#4} & \valuewithsd{#5}{#6}}

\title{From Understanding to Generation: An Efficient Shortcut for Evaluating Language Models}

 \author{Viktor Hangya \and Fabian Küch \and Darina Gold  \\
         Fraunhofer IIS\\
         \texttt{\{first.last\}@iis.fraunhofer.de}}

\begin{document}
\maketitle
\begin{abstract}
Iterative evaluation of LLMs during training is essential to ensure expected capability development, but can be time- and compute-intensive.
While NLU tasks, where the model selects from fixed answer choices, are cheap to evaluate, essential capabilities like reasoning and code generation rely on the more time-consuming NLG (token-by-token generation) format.
In this work, our aim is to decrease the computational burden of NLG benchmarks in order to enable monitoring crucial LLM capabilities during model training.
We reformulate generative tasks into computationally cheaper NLU alternatives.
We test the performance correlation between the original and reformulated tasks using 8 LMs of various sizes and 4 capabilities: mathematical reasoning, code generation, factual knowledge and reading comprehension.
Our results show a strong correlation between task formats, supporting capability assessment via cheaper alternatives and achieving over 35× average reduction in evaluation time.
Our project is available at: \url{https://github.com/Fraunhofer-IIS/EvalShortcut}

\end{abstract}

\section{Introduction}
\label{sec:intro}
The increasing adoption of \glspl{llm} has brought a substantial rise in computational demands—not only for training, but also for evaluating the performance of new models.
Numerous works have explored methods for selecting the best quality training datasets and hyper-parameters based on smaller proxy models,
evaluating them multiple times throughout training \cite{li2024datacomp,grattafiori2024llama,magnusson2025datadecidepredictbestpretraining}.
However, due to model capability and compute limitations they rely on a few benchmarks which support measuring a limited set of model capabilities \cite{penedo2024fineweb,gu-etal-2025-olmes}.

Evaluation benchmarks can be grouped into two types: 
1) \textit{\gls{nlg}} which requires auto-regressive sampling from the model, and 2) \textit{\gls{nlu}} which involves calculating the probability of given answer options \cite{guo2023evaluating,biderman2024lessonstrenchesreproducibleevaluation}.
Assessing a model’s \gls{nlg} abilities is costly, especially when done repeatedly during model training and selection, and thus is often neglected \cite{zhou-etal-2022-deconstructing}.
Yet, many key capabilities, e.g. reasoning and code generation, are only measurable via \gls{nlg} benchmarks.

\begin{table*}[t]
\resizebox{\textwidth}{!}{\begin{tabular}{l|r|lcl|r}
& Step & <Q> := What is gravity? &  & Step output & Final output \\ \hline
\rotatebox{90}{LL} & 1.  & <Q> Downward force on objects.   & $\rightarrow$      & $log(0.9)$  &  Log-likelihood of correct answer: $log(0.9)$   \\ \hline
\multirow{2}{*}{\rotatebox{90}{MC}} & 1.  & <Q> Downward force on objects.    & $\rightarrow$     & \textcolor{red}{$log(0.9)$} & \multirow{2}{*}{Selected answer index: 1.}    \\
 & 2.   & <Q> What makes the sun rise.     & $\rightarrow$      & $log(0.1)$ &    \\ \hline
\multirow{5}{*}{\rotatebox{90}{NLG}} & 1.  &    <Q>  & $\rightarrow$     & Downward & \multirow{5}{*}{Generated answer: Downward force on objects.} \\
 & 2.   & <Q> \textbf{Downward} & $\rightarrow$ &  force &  \\
 &   3.  & <Q> Downward \textbf{force} & $\rightarrow$    & on &  \\
 &  4.   & <Q> Downward force \textbf{on}  & $\rightarrow$   & objects. &  \\
 &  5.   & <Q> Downward force on \textbf{objects.}  & $\rightarrow$       &   [EOS]    & 
\end{tabular}
}
\caption{
Comparison of answers to the same question in the natural language understanding variants (NLU) \textemdash loglikelihood (LL) and multiple-choice (MC)\textemdash and natural language generation (NLG) settings.
LL scores the correct answer to the question (Q), MC selects the most probable answer option, while NLG generates the answer token-by-token.
The column \emph{step} indicates the number of forward model passes needed for the answer calculation.
The example and outputs (e.g., log values) are fictional and serve solely to illustrate the NLU–NLG computation differences.
}
\label{tab:cost_gen_mc_ll}
\end{table*}

Here, we argue for the importance of monitoring the trajectory
of LLM capabilities throughout the (pre-)training process, which is currently reliant on expensive \gls{nlg} benchmarks.

Our goal is to reduce compute requirements for such benchmarks in order to enable a more frequent use.
Towards this end, we propose reformulating expensive \gls{nlg} benchmarks as less-resource intensive \gls{nlu} tasks.
Specifically, we reformulate it to \gls{mc} tasks, where the model has to select the correct answer among answer options, and \gls{ll} tasks where the model calculates the \gls{ll} of the correct answer.
The cost-reduction of \gls{mc} and \gls{ll} over then generative task is shown in Table~\ref{tab:cost_gen_mc_ll}. 
The generative setup completes the response token-by-token. 
The \gls{mc} setup, however, evaluates the model's ability to select the appropriate response from a set of alternatives based on their likelihoods, while the \gls{ll} approach scores only the correct answer.
Despite differing formats, all methods aim to assess the model’s response to the same underlying task.

Since many \gls{nlg} benchmarks do not contain incorrect answer options to questions, we create such options 
using either LLMs or simply picking answers of other samples in a given benchmark randomly.
In contrast, \gls{ll} is simpler to apply, as it
evaluates the likelihood assigned by the model to a given correct answer without the need of an overhead, i.e., the computation of alternative answers. 
As such, it offers further reductions in evaluation costs,
however,
at the expense of evaluation depth, 
since
high-probability incorrect answers are not considered.
\gls{mc} and \gls{ll} evaluation allows quick comparison of model checkpoints and
low-cost training monitoring.
Importantly, our goal is not to replace the NLG format, but to offer a cheaper alternative for monitoring model capabilities during pre-training.

To test this hypothesis, we conduct a study involving a diverse set of benchmarks representing various domains.
We analyze the correlation of 8 open source model performances between the \gls{nlg} and \gls{nlu} task formulations, as well as the correlation of model rankings.
To 
our knowledge, this is the first study to systematically investigate this relationship.
We find a strong correlation, indicating 
the ability to monitor the performance trajectory when training a single model, but also the ability to compare the model in question to baseline LLMs using only the \gls{nlu} formulated benchmarks.
Additionally, we conduct 
ablation studies, e.g.,
pairing \gls{nlg} and \gls{nlu} benchmarks of the same domain but different sources,
finding promising results in these settings, too.
On top of the correlation experiments, we show that besides monitoring the model development trajectory using \gls{nlu} tasks, it is also possible to predict the \gls{nlg} performance of a model by fitting a linear regressor and only occasional \gls{nlg} evaluations.
This optional step balances precise \gls{nlg} performance monitoring and compute resource needs.

\noindent
Our main contributions are threefold:
\begin{itemize}[noitemsep,topsep=0pt,left=0pt]
        \item we propose and apply a methodology for reformulating \gls{nlg} tasks as \gls{nlu} tasks for more efficient benchmark evaluation and broader capability assessment during model selection and training
    \item we show strong correlation between task formulations and analyze various aspects; a very important one being that by reformulating \gls{nlg} tasks we achieve a compute time reduction of 35x on average, reducing runtime from nearly 2 hours to under a minute in the most extreme case.
    
            \item we publicly release our evaluation framework.
\end{itemize}

\section{Related Work}
\label{sec:rel_work}
This section is divided into three subsections, each addressing key components of this work.
In Section~\ref{sec:eval_costs}, we discuss evaluation costs and efforts to reduce them.
In Section~\ref{sec:relation_nlg_nlu}, we review studies on the relation between \gls{nlg} and \gls{nlu} evaluation.
In Section~\ref{sec:task_categorization}, we present knowledge categories used to study LLM generalizability.

\subsection{Evaluation costs}
\label{sec:eval_costs}
Various benchmarks, including \gls{nlg} with many samples, have been proposed to evaluate \glspl{llm}, but evaluating large models on them can be costly.
For example, \citet{liang2023holistic} introduce the HELM benchmark which could cost over 4K GPU hours to evaluate a single \gls{llm}, not to mention monitoring the performance over the course of the training process \cite{biderman2023pythia,liu2024llm}.
Recent studies have proposed various strategies to address
reducing the evaluation costs.
While \citet{perlitz2023efficient} and \citet{polo2024tinybenchmarks} reduce the number of evaluation items using subsampling,
\citet{kuramoto2025predicting} predict fine-tuning outcomes on large datasets using results from smaller-scale experiments.
Compressed benchmarks may reduce evaluation diversity and underestimate model limitations by missing out on key evaluation samples.
While \citet{polo2024tinybenchmarks} retain key samples via an information-theoretic approach, their approach requires sample-level correctness results of a set of LLMs making its application difficult.
In contrast, we reformulate \gls{nlg} tasks to \gls{nlu} independently from the target LLM, reducing costs while retaining all samples.

\subsection{Linking \gls{nlg} and \gls{nlu}}
\label{sec:relation_nlg_nlu}
Several studies attempt to use \gls{nlu} tasks to assess \gls{nlg} and reduce costs \cite{gsm8k-mc,khashabi2020unifiedqa,li2023seed,myrzakhan2024open}, but mainly target single tasks, such as mathematical reasoning \cite{gsm8k-mc} or \gls{qa}
\cite{khashabi2020unifiedqa}.

Most similar to our work, \citet{gsm8k-mc} compare \gls{mc} and \gls{nlg} versions of math and coding benchmarks, showing evaluation time can be reduced by up to 30×.
Our work however, differs in multiple aspects, such as we explore both \gls{mc} and \gls{ll} reformulations, we explore and use multiple approaches to build answer options for \gls{mc} tasks including a lightweight method which does not rely on LLM-based generation, we test our approach on a diverse set of capabilities and present the first systematic analysis of NLG–MC benchmark pairings showing consistently positive correlation.
Furthermore, we show positive correlation on all of them in contrast to \citet{gsm8k-mc} who showed mixed results.

\citet{khashabi2020unifiedqa} introduce a single \gls{qa} model that performs well across multiple formats, including \gls{mc} and generative. 
Their study shows that knowledge learned from \gls{mc} tasks transfers effectively to generative \gls{qa}.
However, their approach requires model adaptations, while our approach requires reformulating a given benchmark once which is then applicable to any \gls{llm}.

\citet{li2023seed} propose a method to create \gls{mc} benchmarks 
for evaluating
multimodal \glspl{llm}. 
They do so by prompting language models with visual content extracted from images or videos
to generate one question and four answer options (one correct), effectively converting open-ended understanding into a structured \gls{mc} format.
We follow a similar approach, to generate distractor answer options in our \gls{mc} reformulations.

Finally, \citet{myrzakhan2024open} convert \gls{mc} benchmarks into open-ended 
formats to better assess generative abilities and reduce biases such as guessing. 
They 
find that open-style questions 
score lower, suggesting \gls{mc} questions may overestimate model ability. 
While \citet{myrzakhan2024open} did not report the correlation between \gls{mc} and \gls{nlg} formats, we computed Pearson correlation
showing a positive relation between task formulations.

\subsection{Task categorization for LLM evaluation}
\label{sec:task_categorization}
As noted, prior \gls{nlg}–\gls{nlu} studies focus on specific tasks; we aim to generalize across multiple capabilities.
\citet{wang2024ubench} evaluate LLMs' reliability in \gls{mc} questions across knowledge, language, understanding, and reasoning tasks.
\citet{wang2024fac} break down evaluation in four different neural capabilities: linguistic knowledge, formal knowledge, world modeling, and social modeling.
In this work we consider four \gls{llm} capabilities: mathematical reasoning, factual knowledge, reading comprehension and code generation.

\section{Experiment Setup}
\label{sec:ex_setup}
In this section, we describe our experiment setup, its components and motivate our choices.
This paper investigates the hypothesis that \gls{ll} and \gls{mc} formulations can serve as effective proxies for assessing generative model capability development over the course of training, while significantly reducing the computational cost of evaluation.
To evaluate this hypothesis, we pair different formulations of the same or related benchmarks, analyze the correlation between the \gls{nlg} and \gls{nlu} versions, and compare their evaluation runtimes.

\subsection{Evaluation variants}
We use three different variants of each benchmark: the original \gls{nlg} version, the \gls{mc} version and the \gls{ll} version.
In the \gls{nlg} variant, the model receives a question and must produce an answer as free text.
In the \gls{ll} and \gls{mc} variants, it is given the question along with either a single correct answer option (\gls{ll}) or multiple (a correct and multiple incorrect) answer options (\gls{mc}) and calculates the probabilities assigned to them.
In Table~\ref{tab:cost_gen_mc_ll}, we illustrate these settings.

To judge the accuracy of the outputs different metrics are considered depending on the task. For the generative formulation, exact (token-by-token) or proximity (e.g. BLEU)
matching with the gold answer are used.
For the \gls{mc} formulation, the model is considered to have selected the correct answer if it was the one with the highest probability.
Since \gls{ll} only scores the correct answers, we calculate the average log-probability ($mean_{i=1..N}log\hat{P}(y_i|x_i)$\footnote{We take (character) length normalized values: $log\hat{P}(y_i|x_i) = logP(y_i|x_i)/|y_i|$.}) of correct answers ($y_i$) of questions ($x_i$) in a given benchmark
and we expect that it correlates with quality of the model's capability without calculating an exact correctness value. We detail calculating correlation in Section~\ref{sec:ex_setup_metrics}.

As shown in Equation~\ref{eq:time_complexity},
log-likelihood evaluates the probability of a fixed target sequence in a single forward step,
multiple choice evaluates the given (\( K \)) candidate sequences in one forward step each, while
generative decoding 
generates one token in each forward step over \( T \) time steps (maximum output tokens).
For further details about the exact calculations we refer to \citet{eval-harness}.

\vspace{-18pt}
\begin{equation}
\small
\text{Compute Complexity} =
\begin{cases}
\mathcal{O}(1)           & \text{(\gls{ll})} \\
\mathcal{O}(K)           & \text{(\gls{mc})} \\
\mathcal{O}(T) & \text{(\gls{nlg})}
\end{cases}
\quad \text{where } K \ll T
\label{eq:time_complexity}
\end{equation}

\subsection{Metrics}
\label{sec:ex_setup_metrics}
To evaluate how indicative \gls{nlu} task formulations are on the generative performance, we use both intra- and cross-model metrics.

The goal of intra-model metrics is to measure how similarly the performance of \gls{nlu} and \gls{nlg} versions of the same task change over the training course of a model.
See Figure~\ref{fig:corr_gen_vs_mc} for a visualization.
We calculate Pearson correlation between the \gls{nlu} and \gls{nlg} results of the intermediate model checkpoints over the course of training.
We define $P_{macro}$ and $P_{micro}$ as macro- and micro-averaged Pearson correlation across models respectively, i.e., calculating the correlation values for each of our 8 considered models separately followed by averaging the model-wise correlation values, and calculating the correlation using data points of all models jointly.
High correlation values indicate that improvements or decrease of model performance on an \gls{nlu} task during the training process means performance change in the same direction in the \gls{nlg} format as well.

In contrast, the goal of cross-model metrics is to measure the consistency of the ranking of our 8 models in the different task formulations.
For this, we take the model rankings based on averaged performance over all intermediate checkpoints per model and calculate Spearman correlation.
High coefficients indicate the possibility of comparing the model of interest to other models using the \gls{nlu} task formulation instead of running the costly \gls{nlg} version.

\subsection{Benchmarks}
To show the broad applicability of our approach, we use four distinct knowledge domains. This section discusses existing benchmarks; benchmark reformulations are detailed in Section~\ref{sec:distractor_creation}.
Examples for all benchmarks discussed herein as well as their extensions are shown in Table~\ref{tab:example_all_benchmarks} in the appendix.

\paragraph{\gls{nlg} Benchmarks}
We leverage GSM8K \cite{gsm8k} for mathematical reasoning, HumanEvalPack \cite{muennighoff2023octopack}\footnote{This is the extension of HumanEval \cite{human-eval} from Python to 5 more code languages: C++, Go, Java, JavaScript and Rust} for code generation (referenced as HumanEval further on), TriviaQA \cite{triviaqa} for factual knowledge and SQuAD 2.0\footnote{As we use only this version of \textit{SQuAD}, it will be referenced without the version further on.} \cite{squad} for reading comprehension.

\paragraph{\gls{nlu} Benchmarks}
We discuss off-the-shelf \gls{nlu} benchmarks in this section, while the reformulation of \gls{nlg} tasks as \gls{mc} is covered in Secion~\ref{sec:distractor_creation}.
\citet{gsm8k-mc} created a \gls{mc} version of GSM8K by using incorrect answers of LLMs, and only the numeric value (see Table~\ref{tab:example_all_benchmarks}), to a given question as \gls{mc} options. 
We use this version of GSM8K, besides our own, to represent \gls{mc} tasks for mathematical reasoning.
Additionally, we run ablation studies where we test cross-benchmark pairings, see Section~\ref{sec:distractor_creation}.
For these experiments, we use MMLU \cite{mmlu} for factual knowledge
and BoolQ \cite{boolq} for reading comprehension.

\subsection{Multiple choice distractor creation}
\label{sec:distractor_creation}
Since most \gls{nlg} benchmarks only provide the correct answers, which is already enough for \gls{ll}, we had to create incorrect answer options in order to reformulate them as \gls{mc} tasks.
We focus on the creation of \gls{mc} benchmark variants using existing \gls{nlg} benchmarks, as done by \citet{gsm8k-mc}.
We created both random and smart 
negative answer options (3 if not stated otherwise) which we discuss next.
Table~\ref{tab:ex_random_smart_distractors} shows exemplary random and smart distractors to the same question.

\begin{table}[t]
\begin{tabular}{p{2.5cm}|p{4.5cm}}
\hline
Question           & Which was the first European country to abolish capital punishment? \\ \hline
Correct answer     & Norway                                                              \\ \hline
Random \\ distractors & {[}Chicago Bears, Ballet, 6{] }      \\        \hline
Smart \\ distractors  & {[}Germany, Italy, Poland{]}  \\ 
\hline
\end{tabular}
\caption{Example for random and smart distractors in TriviaQA.}
\label{tab:ex_random_smart_distractors}
\end{table}

\paragraph{Random distractors} 
For GSM8K, TriviaQA, and HumanEvalPack, we created random distractors\footnote{
We only test a single random seed for efficiency reasons.}
by using answers of other questions within the same benchmark.
For SQuAD, we did it similarly, but used only answers of questions that have the same context as the question at hand.\footnote{SQuAD questions are created based on Wikipedia articles. Each article yields multiple questions. Table~\ref{tab:example_all_benchmarks} shows two questions based on the same article.}
This ensured that they were formally valid, but semantically incorrect.
However, we hypothesized that models might disregard the incorrect answers not because they knew the correct one, but because the incorrect options were implausible or clearly irrelevant to the question, e.g. as shown in Table~\ref{tab:ex_random_smart_distractors}, the question is looking for a \textit{European country} and none of the random distractors is a location.
Hence, we also created \textit{smart distractors}.

\paragraph{Smart distractors}
\label{sec:ex_setup_more_versions}
To generate three plausible incorrect answers\textemdash \textit{smart distractors}, we prompted a Meta-Llama-3.1-70B-Instruct-GPTQ-INT4 model (using a benchmark-specific 5-shot prompt; see Appendix~\ref{sec:app:distractor_prompt}) for two \gls{nlg} benchmarks, namely TriviaQA and GSM8K.
Table~\ref{tab:ex_random_smart_distractors} illustrates how smart distractors, unlike \textit{random distractors}, match the semantic context (e.g. European countries). 
To assess the plausibility of the automatically generated smart distractors, one author manually reviewed 50 random question samples and their corresponding distractors for each benchmark. 
Out of the overall 100 questions, one distractor was a correct answer, and in eight cases, two of the distractors were repeated, leaving only two distinct wrong answers.
After we developed our initial prompt for the first task (Appendix~\ref{sec:app:distractor_prompt}), adapting it to the next took under 30 minutes.
Depending on the task, running the distractor generation took between 16-24 hours, including some manual checks and fixing JSON parsing errors due to incorrect LLM outputs.
We believe this demonstrates that our prompt-based approach is an effective and fast method for generating smart distractors.
For SQuAD, we omitted smart distractors since random distractors already share context with the question. 
For HumanEvalPack, as smart distractors we use the slightly altered incorrect code snippets provided by the dataset for the code debugging task, resulting in two \gls{mc} options instead of four.

\paragraph{Cross-benchmark pairing}

On top of converting \gls{nlg} benchmarks to \gls{mc} using either random or smart distractor generation, we experiment with pairing \gls{nlg} benchmarks with off-the-shelf \gls{mc} datasets targeting similar capabilities.
However, we note that these secondary experiments serve only as a scientific exploration to test performance correlation on related but different datasets, since the above mentioned distractor generation is trivial and not all LLM capabilities have off-the-shelf \gls{nlg} and \gls{mc} benchmarks available.
To the best of our knowledge, this is the first work to conduct such a pairing analysis between \gls{nlg} and \gls{mc} benchmarks.
We found pairings for two of our \gls{nlg} datasets.
To represent \textit{factual knowledge}, we use MMLU as the \gls{mc} version of TriviaQA.
Similarly, to represent \textit{reading comprehension}, we use BoolQ as the \gls{mc} version of SQuAD.

\subsection{Models}
\label{sec:ex_setup_models}
We selected open base models of varying sizes and training stages, enabling evaluation across a range of performance levels and intermediate checkpoints, which aligns with our motivation to efficiently monitor LLM capabilities during pretraining.
We use the Pythia\footnote{We used the deduplicated version.} 1B, 2.8B and 6.9B \cite{biderman2023pythia}, Amber 7B \cite{liu2024llm}, OLMo 1B and 7B \cite{groeneveld-etal-2024-olmo}, and OLMo-2 7B models \cite{olmo20242}, as well as the code-specific Crystal 7B model \cite{liu2024llm}.
We did not include larger LLMs due to compute restrictions.
We evaluated intermediate model checkpoints at 20B, 40B, 60B, 100B, 300B, 1T, 1.3T, 2T, 3T and 4T tokens.
Note that not all models were trained up to 4T tokens.
Please see Table~\ref{tab:model_ckpts} for more details.
We present results based on the amount of FLOPS used to produce a given checkpoint in figures~\ref{fig:corr_gen_vs_mc} and~\ref{fig:pred}. 
We follow the standard approach to estimate FLOPS based on model parameter\footnote{We exclude embedding parameters as suggested by~\cite{kaplan2020scaling}.
} (N) and token (D) counts: $FLOPS = 6ND$ \cite{kaplan2020scaling}.

\subsection{Technical details}

For our experiments, we used the \emph{lm-eval-harness}~\cite{eval-harness}, due to its wide adoption, extensibility, and
support for a broad range of benchmarks.
We extend this framework by adding setups of our new \gls{ll} and \gls{mc} formulations of the mentioned benchmarks.
The only exception is HumanEval for which we use the \emph{bigcode-evaluation-harness}~\cite{bigcode-evaluation-harness} as it supports safe code execution.
Still, we add its \gls{nlu} formulations to the lm-eval-harness as these formats do not need code execution.
We use default parameters of the evaluation frameworks, except that we use a 0-shot setup for all tasks other than all variants of GSM8K, for which we use 5-shot examples.
We follow the suggestions of OLMES~\cite{gu2024olmes} when evaluating \gls{mc} benchmarks, i.e., we use the completion (cloze) formatting, where models score answer options instead of answer labels (A, B, C, D, etc.), and we use length normalized probability values to select the final answer for accuracy calculation.
For further details we refer to the above mentioned papers.

\begin{table}[t]
\centering
\resizebox{\columnwidth}{!}{\begin{tabular}{ll|rrr}
NLG & NLU & $P_{macro}$ & $P_{micro}$ & Spearman \\
\hline\hline
  \multirow{3}{*}{\rotatebox{90}{GSM8K}}                     & \gls{mc}           & \tablevaluewithsdsix{0.75}{0.12}{0.52}{0.00}{0.76}{0.03} \\
                                                             & \gls{mc}$_{rnd}$   & \tablevaluewithsdsix{0.76}{0.11}{0.57}{0.00}{0.76}{0.03} \\
                                                             & \gls{ll}           & \tablevaluewithsdsix{0.79}{0.09}{0.56}{0.00}{0.81}{0.01} \\
	\hline
  \multirow{3}{*}{\rotatebox{90}{Trivia}}                    & \gls{mc}           & \tablevaluewithsdsix{0.90}{0.03}{0.94}{0.00}{0.86}{0.01} \\
                                                             & \gls{mc}$_{rnd}$   & \tablevaluewithsdsix{0.91}{0.02}{0.88}{0.00}{0.98}{0.00} \\
                                                             & \gls{ll}           & \tablevaluewithsdsix{0.90}{0.03}{0.69}{0.00}{0.81}{0.01} \\
	\hline
  \multirow{2}{*}{\rotatebox{90}{\parbox{0.9cm}{SQu-\\AD}}}                     & \gls{mc}$_{rnd}$   & \tablevaluewithsdsix{0.90}{0.03}{0.88}{0.00}{0.93}{0.00} \\
                                                             & \gls{ll}           & \tablevaluewithsdsix{0.65}{0.15}{0.85}{0.00}{0.69}{0.06} \\
	\hline
  \multirow{3}{*}{\rotatebox{90}{\parbox{1.1cm}{Human\\Eval}}} & \gls{mc}           & \tablevaluewithsdsix{0.83}{0.09}{0.79}{0.00}{0.81}{0.02} \\
	                                                           & \gls{mc}$_{rnd}$   & \tablevaluewithsdsix{0.85}{0.07}{0.75}{0.00}{0.81}{0.02} \\
	                                                           & \gls{ll}           & \tablevaluewithsdsix{0.86}{0.07}{0.73}{0.00}{0.79}{0.03} \\
\end{tabular}}
\caption{
Correlation statistics of \gls{nlg} tasks and their various reformulated formats.
We present p-values in parentheses.
HumanEval results averaged over the 6 coding languages.
\gls{mc} stands for the multiple-choice with smart distractors,
\gls{mc}$_{rnd}$ random distractors,
\gls{ll} is the log-likelihood formulation.
}
\label{tab:corr_gen_vs_mc}
\end{table}

\section{Results}
\label{sec:discussion}
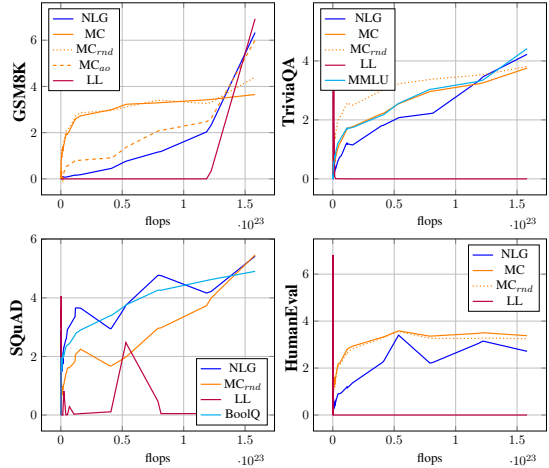
\begin{figure}[t]
    \centering
    \resizebox{.45\textwidth}{!}{    \begin{tikzpicture}
            \begin{axis}[
            xlabel={flops},
                        title={\textbf{GSM8K}},
            title style={
                rotate=90,
                at={(-0.05,0.5)},                  anchor=south,
                font=\Large
            },
            grid=major,
            legend pos=north west,          ]
        \addplot[blue, line width=1pt,] table [x=flops, y=gsm8k, col sep=comma] {flop_task_st.csv};
        \addlegendentry{NLG};
        \addplot[orange, line width=1pt,] table [x=flops, y=gsm8k_mc_cloze_cot, col sep=comma] {flop_task_st.csv};
        \addlegendentry{\gls{mc}};
        \addplot[orange, line width=1pt, dotted] table [x=flops, y=gsm8k_mc_cloze_rand, col sep=comma] {flop_task_st.csv};
        \addlegendentry{\gls{mc}$_{rnd}$};
        \addplot[orange, line width=1pt, dashed] table [x=flops, y=gsm8k_mc_cloze, col sep=comma] {flop_task_st.csv};
        \addlegendentry{\gls{mc}$_{ao}$};
        \addplot[purple, line width=1pt,] table [x=flops, y=gsm8k_ll_correct_prob_norm, col sep=comma] {flop_task_st.csv};
        \addlegendentry{\gls{ll}};
                        \end{axis}
        
            \begin{axis}[
            xlabel={flops},
                        title={\textbf{TriviaQA}},
            title style={
                rotate=90,
                at={(-0.05,0.5)},                  anchor=south,
                font=\Large
            },
            grid=major,
            legend pos=south east,                          at={(8cm, 0)},
        ]
        \addplot[blue, line width=1pt,] table [x=flops, y=triviaqa, col sep=comma] {flop_task_st.csv};
        \addlegendentry{NLG};
        \addplot[orange, line width=1pt,] table [x=flops, y=triviaqa_mc_cloze, col sep=comma] {flop_task_st.csv};
        \addlegendentry{\gls{mc}};
        \addplot[orange, line width=1pt, dotted] table [x=flops, y=triviaqa_mc_cloze_rand, col sep=comma] {flop_task_st.csv};
        \addlegendentry{\gls{mc}$_{rnd}$};
        \addplot[purple, line width=1pt,] table [x=flops, y=triviaqa_ll_correct_prob_norm, col sep=comma] {flop_task_st.csv};
        \addlegendentry{\gls{ll}};
        \addplot[cyan, line width=1pt,] table [x=flops, y=mmlu_continuation, col sep=comma] {flop_task_st.csv};
        \addlegendentry{MMLU};
        \end{axis}

            \begin{axis}[
            xlabel={flops},
                        title={\textbf{SQuAD}},
            title style={
                rotate=90,
                at={(-0.05,0.5)},                  anchor=south,
                font=\Large
            },
            grid=major,
            legend pos=south east,                          at={(0cm, -7cm)},
        ]
        \addplot[blue, line width=1pt,] table [x=flops, y=squadv2, col sep=comma] {flop_task_st.csv};
        \addlegendentry{NLG};
        \addplot[orange, line width=1pt,] table [x=flops, y=squadv2_mc_4, col sep=comma] {flop_task_st.csv};
        \addlegendentry{\gls{mc}$_{rnd}$};
        \addplot[purple, line width=1pt,] table [x=flops, y=squadv2_ll_correct_prob_norm, col sep=comma] {flop_task_st.csv};
        \addlegendentry{\gls{ll}};
        \addplot[cyan, line width=1pt,] table [x=flops, y=boolq, col sep=comma] {flop_task_st.csv};
        \addlegendentry{BoolQ};
        \end{axis}

            \begin{axis}[
            xlabel={flops},
                        title={\textbf{HumanEval}},
            title style={
                rotate=90,
                at={(-0.05,0.5)},                  anchor=south,
                font=\Large
            },
            grid=major,
            legend pos=south east,                          at={(8cm,-7cm)}
        ]
        \addplot[blue, line width=1pt,] table [x=flops, y=humanevalpack_gen_avg, col sep=comma] {flop_task_st.csv};
        \addlegendentry{NLG};
        \addplot[orange, line width=1pt,] table [x=flops, y=humanevalpack_cloze_debug_avg, col sep=comma] {flop_task_st.csv};
        \addlegendentry{\gls{mc}};
        \addplot[orange, line width=1pt, dotted] table [x=flops, y=humanevalpack_cloze_avg, col sep=comma] {flop_task_st.csv};
        \addlegendentry{\gls{mc}$_{rnd}$};
        \addplot[purple, line width=1pt,] table [x=flops, y=humanevalpack_ll_correct_prob_norm_avg, col sep=comma] {flop_task_st.csv};
        \addlegendentry{\gls{ll}};
        \end{axis}
    \end{tikzpicture}
    }
    \caption{
        Average performance across models per task formulation at fixed compute (flops).
    Task formulation performances are standardized to bring different formulations to a more visually comparable scale.
    Our expectation is that \gls{nlg} and its reformulations have similar developments over time, i.e., high Pearson correlation.
                            }
    \label{fig:corr_gen_vs_mc}
\end{figure}

Table~\ref{tab:corr_gen_vs_mc} shows our main results of the four considered task categories.
Each row shows the correlation coefficients for the results of the given task in the \gls{nlg} and indicated reformulated setting.
We present averaged code results on 6 coding languages (cpp, go, java, js, python and rust).\footnote{Detailed results can be found in Table~\ref{tab:all_results}.}
As introduced, we have three reformulated versions of the \gls{nlg} tasks: \gls{mc} and \gls{mc}$_{rnd}$ indicate results for the multiple-choice reformulation with smart and random distractors respectively, while \gls{ll} represents the log-likelihood reformulation.
Additionally, Figure~\ref{fig:corr_gen_vs_mc} shows task performance curves over the course of training, averaged across models and standardized for each task formulation to bring different formulations to more visually comparable scales.

Overall, Table~\ref{tab:corr_gen_vs_mc} demonstrates a strong correlation between the \gls{nlg} and both \gls{mc} task variants, as well as between \gls{nlg} and \gls{ll}, in all three metrics.
This trend is also supported in Figure~\ref{fig:corr_gen_vs_mc}, showing similar performance shifts on the x-axis between the \gls{nlg} and \gls{nlu} formulations.
These results suggest that the \gls{nlu} reformulated tasks are effective indicators during model training for the generative performance and can reduce compute.

Interestingly, \gls{mc} and \gls{mc}$_{rnd}$ perform on par with each other.
We compared the two formulations on small (1B and 2.8B) and large (6.9B and 7B)
models, and found that in case of small models \gls{mc}$_{rnd}$ performs clearly better
than \gls{mc}
($P_{macro}$: +0.05, $P_{micro}$: +0.14, Spearman: +0.52),
while in case of large models
we did not find a clear difference
($P_{macro}$: -0.02, $P_{micro}$: 0.00, Spearman: 0.13 when comparing \gls{mc}$_{rnd}$ to \gls{mc}).
Our conjecture is that less capable models tend to output answers unrelated to the question, e.g., completely wrong GSM8K reasoning sequence, thus can be misled by random answer options more easily, while better quality models understand the context of questions more precisely, thus are less sensitive to random distractors.

Although \gls{ll} performs on par with \gls{mc} and \gls{mc}$_{rnd}$, the correlation values are slightly lower on average, especially on TriviaQA, SQuAD and HumanEval.
On the contrary \gls{ll} performs slightly better on GSM8K.
These results are in line with the findings of \citet{schaeffer2024why}, who found that it is important to consider the probabilities of a few negative answer options when predicting scaling behavior of LLMs.
In addition, our results show that considering negative options in the form of the \gls{mc} task formulation is beneficial also for monitoring the generative performance of a given model.
Although the effort of creating smart and random distractors for the \gls{mc} formulations is minimal, one can sacrifice a slight performance for the simplicity of \gls{ll} which only needs the correct answer options.
The better performance of \gls{ll} on GSM8K 
is likely attributable to the \gls{CoT} reasoning steps in the outputs which makes the log-likelihood calculation more reliable.
As discussed in section~\ref{sec:ablation}, when we omit the reasoning steps from the scored output the performance drops significantly.
This could indicate the need for more difficult distractor options.

As discussed before, the goal of cross-model metric (Spearman) is to compare different models using the cheaper reformulated tasks.
In contrast, the goal of intra-model metrics ($P_{macro}$ and $P_{micro}$) is to show whether reformulated tasks can be used to monitor the \gls{nlg} performance of a single model during training.
We found similarly strong correlation values for both categories, showing their usefulness in both scenarios.
Comparing $P_{macro}$ and $P_{micro}$, we found stronger coefficients for $P_{macro}$
indicating that the scaling factor between the \gls{nlg} and the reformulated task performances slightly varies from model to model.

\begin{table}[t]
\centering
\resizebox{\columnwidth}{!}{\begin{tabular}{ll|rrr}
NLG & NLU & $P_{macro}$ & $P_{micro}$ & Spearman \\
\hline\hline
  \multirow{4}{*}{\rotatebox{90}{GSM8K}}                     & \gls{mc}$_{rnd}$   & \tablevaluewithsdsix{0.76}{0.11}{0.57}{0.00}{0.76}{0.03} \\
                                                             & \gls{mc}$_{ao}$    & \tablevaluewithsdsix{0.38}{0.26}{0.90}{0.00}{0.88}{0.00} \\
                                                             & \gls{ll}           & \tablevaluewithsdsix{0.79}{0.09}{0.56}{0.00}{0.81}{0.01} \\
                                                             & \gls{ll}$_{ao}$    & \tablevaluewithsdsix{0.22}{0.38}{-0.04}{0.80}{-0.07}{0.87} \\
	\hline
  \multirow{2}{*}{\rotatebox{90}{Trivia}}                    & \gls{mc}$_{rnd}$   & \tablevaluewithsdsix{0.91}{0.02}{0.88}{0.00}{0.98}{0.00} \\
                                                             & MMLU               & \tablevaluewithsdsix{0.89}{0.04}{0.94}{0.00}{0.95}{0.00} \\
	\hline
  \multirow{3}{*}{\rotatebox{90}{SQuAD}}                     & \gls{mc}$_{rnd}$   & \tablevaluewithsdsix{0.90}{0.03}{0.88}{0.00}{0.93}{0.00} \\
                                                             & \gls{mc}$_{rnd^*}$ & \tablevaluewithsdsix{0.90}{0.03}{0.84}{0.00}{0.93}{0.00} \\
                                                             & BoolQ              & \tablevaluewithsdsix{0.78}{0.11}{0.24}{0.09}{0.45}{0.26} \\

\end{tabular}}
\caption{
Correlation statistics of various additional tests.
For GSM8K we tested setups where only the final answer has to be scored by the model (*$_{ao}$) as proposed by \cite{gsm8k-mc}.
In case of TriviaQA and SQuAD, we tested  cross-benchmark pairings: MMLU and BoolQ respectively.
Additionally, in \gls{mc}$_{rnd^*}$ we used more than 4 answer options for SQuAD.
}
\label{tab:custom_results}
\end{table}

\subsection{Ablations}
\label{sec:ablation}

In addition to the above experiments, we evaluated further task formulations to have a better understanding of the important factors which we present in Table~\ref{tab:custom_results}.
For GSM8K, as discussed above, having \gls{CoT} reasoning steps in the scored outputs is an important factor for the correlation values.
Based on the work of \citet{gsm8k-mc}, we evaluate the \gls{mc} and \gls{ll} formulations of GSM8K which only include the final numeric \underline{a}nswer \underline{o}nly in the output (\gls{mc}$_{ao}$ and \gls{ll}$_{ao}$ respectively).
The reasoning steps are crucial for the \gls{ll} formulation as the correlation coefficients of all three metrics drop significantly when removing them.
In contrast, providing only the final answer has the opposite effect in case of \gls{mc}.
Figure~\ref{fig:corr_gen_vs_mc} shows that in contrast to \gls{mc}, \gls{mc}$_{ao}$ reflects the rapid performance increase at later stages more precisely, while \gls{mc} seems too easy for the models, i.e, the performance rapidly increases at the early training stages and slows down later on.
This indicates that harder distractors might be needed for this task.

For TriviaQA and SQuAD we tested cross-benchmark pairing by leveraging the  MMLU and BoolQ datasets.
Surprisingly, we found strong correlation values using MMLU, even outperforming our \gls{mc}$_{rnd}$ formatted TriviaQA version.
A possible explanation for this is the longer answer options in MMLU compared to the short options in TriviaQA which could make MMLU more reliable in the \gls{mc} format.
In contrast, BoolQ does not perform as well as SQuAD \gls{mc}$_{rnd}$, although it still has a decent correlation ($P_{macro}$).
As can be seen in Figure~\ref{fig:corr_gen_vs_mc}, BoolQ nicely reflects the \gls{nlg} performance increase of SQuAD on average,
however, it does not follow the curve as closely.
This aligns well with the findings with MMLU, as BoolQ has only short \emph{Yes} and \emph{No} answer options which could make it less reliable.
These findings also align with \citet{xiao2024densinglawllms}, who argue for loss calculation on longer outputs when finding the effective parameter size of LLMs, and opt for LLM based \gls{CoT} generation in case of tasks with short outputs.
We leave this for future work, as reasoning generation for factual tasks is beyond our scope.

Finally, since multiple questions were derived from a given Wikipedia context in SQuAD, we have the options to create more than 3 related negative answer options for each question.
In \gls{mc}$_{rnd^*}$, we use 6 negative answer options on average ($\pm2.22$; at most 16) depending on the source context.
We find minimal difference from \gls{mc}$_{rnd}$, highlighting the robustness of 4 overall answer options that is frequently used for \gls{mc} tasks.

\subsection{Runtimes}
\label{sec:runtime}

The main motivation for reformulating \gls{nlg} tasks is to reduce compute time requirements.
In Table~\ref{tab:runtime} we present task runtimes in minutes for three Pythia\footnote{We only consider Pythia models here, as they feature the same architecture across a wide range of sizes.} model sizes: 1B, 2.8B and 6.9B.
Overall, the \gls{mc} and \gls{ll} reformulations bring significant runtime reductions on our benchmarks (2x--176x averaged across model sizes) compared to their \gls{nlg} counterparts, becoming more and more efficient as model size grows.
The most significant improvement was achieved for code generation, where the runtime was reduced from nearly 2 hours\footnote{Due to the length of the outputs. We use the suggested maximum generation length of 2048.} to under a minute.
As expected \gls{ll} is more efficient than \gls{mc} since it has only one answer option to score.
These improvements significantly reduce runtime costs during model evaluation, while causing no additional effort when reformulating \gls{nlg} tasks to \gls{ll} and only minimal costs when generating random or smart distractors for the \gls{mc} tasks.

\begin{table}[t]
\centering
\resizebox{\columnwidth}{!}{\begin{tabular}{ll|rrr|r}
                        &      & 1B    & 3B     & 7B     & Avg.\textcolor{gray}{\scriptsize(Imp.)}\hspace{3mm}   \\ \hline
\multirow{3}{*}{\rotatebox{90}{GSM8K}}  & NLG & 4.77  & 12.55  & 27.08  & 14.80\hspace{9mm}  \\
                        & MC   & 2.45  & 5.88   & 12.90  & 7.08\textcolor{gray}{\scriptsize(2.1x)}\hspace{3mm}   \\
                        & LL   & 1.05  & 2.08   & 3.93   & 2.36\textcolor{gray}{\scriptsize(6.3x)}\hspace{3mm}   \\ \hline
\multirow{3}{*}{\rotatebox{90}{Trivia}} & NLG & 11.23 & 32.40  & 47.90  & 30.51\hspace{9mm}  \\
                        & MC   & 1.13  & 1.93   & 3.75   & 2.27\textcolor{gray}{\scriptsize(13.4x)}\hspace{2mm}   \\
                        & LL   & 1.22  & 1.42   & 1.85   & 1.49\textcolor{gray}{\scriptsize(20.5x)}\hspace{2mm}   \\ \hline
\multirow{3}{*}{\rotatebox{90}{SQuAD}}  & NLG & 17.70 & 53.08  & 144.27 & 71.68\hspace{9mm}  \\
                        & MC   & 4.68  & 10.57  & 23.50  & 12.92\textcolor{gray}{\scriptsize(5.5)}\hspace{4.5mm}  \\
                        & LL   & 1.77  & 3.53   & 7.03   & 4.11\textcolor{gray}{\scriptsize(17.4x)}\hspace{2mm}   \\ \hline
\multirow{3}{*}{\rotatebox{90}{\parbox{1cm}{Human\\Eval}}}   & NLG & 66.75 & 138.89 & 139.11 & 114.92\hspace{9mm} \\
                        & MC   & 0.54  & 0.83   & 1.26   & 0.88\textcolor{gray}{\scriptsize(130.6x)}   \\
                        & LL   & 0.48  & 0.66   & 0.83   & 0.65\textcolor{gray}{\scriptsize(176.3x)}  
\end{tabular}}
\caption{
Task runtimes of a single model checkpoint in minutes on a single Nvidia RTX 6000.
For consistency, we present Pythia models only at three different model sizes (1B, 3B and 7B).
We present averaged runtime over the three model sizes in columns \emph{Avg.} as well as speed improvements compared to the generative formulation in parenthesis.
}
\label{tab:runtime}
\end{table}

\begin{table}[t]
\centering
\resizebox{.75\columnwidth}{!}{\begin{tabular}{ll|rr}
	NLG                                               & NLU              & Err.  & Spearman \\
	\hline\hline
	\multirow{3}{*}{\rotatebox{90}{GSM8K}}            & \gls{mc}         & 0.031 & 0.48     \\
	                                                  & \gls{mc}$_{rnd}$ & 0.038 & 0.43     \\ 
	                                                  & \gls{ll}         & 0.021 & 0.62     \\ \hline
	\multirow{3}{*}{\rotatebox{90}{Trivia}}           & \gls{mc}         & 0.054 & 0.98     \\
	                                                  & \gls{mc}$_{rnd}$ & 0.047 & 1.00     \\ 
	                                                  & \gls{ll}         & 0.057 & 0.92     \\ \hline
	\multirow{2}{*}{\rotatebox{90}{\parbox{.9cm}{SQu-AD}}} & \gls{mc}$_{rnd}$ & 0.046 & 0.93  \\ 	                                                  & \gls{ll}         & 0.064 & 0.95     \\ \hline
	\multirow{3}{*}{\rotatebox{90}{\parbox{1.1cm}{Human\\Eval}}}  & \gls{mc} & 0.025 & 0.87  \\
	                                                  & \gls{mc}$_{rnd}$ & 0.030 & 0.79     \\
	                                                  & \gls{ll}         & 0.021 & 0.92     \\
				\end{tabular}}
\caption{
Results of predicting the \gls{nlg} performance based on \gls{nlu}.
Err. represents the absolute error between the true and predicted scores, while Spearman indicates model ranking similarity.
}
\label{tab:pred}
\end{table}

\subsection{Predicting \gls{nlg} performance}
\label{sec:mc_to_gen}
As mentioned before, we do not aim to completely eliminate \gls{nlg} tasks but to reduce compute needs during model training by relying on the \gls{nlu} reformulations.
Running occasional \gls{nlg} evaluations is still beneficial as it gives exact model performance on the \gls{nlg} formulations.
In order to further reduce the frequency of such occasional \gls{nlg} evaluations, in this section we aim at predicting the \gls{nlg} performance of a model by training linear regression models.
More precisely, to predict the \gls{nlg} performance of a given model at timestep $i$, we leverage both \gls{nlg} and \gls{nlu} performance scores at timesteps $i-1$, $i-2$ and $i-3$ as training data and \gls{nlu} performance at timestep $i$ as input for the prediction.
Table~\ref{tab:pred} presents absolute prediction error (Err.), and similarly as before, Spearman correlation coefficients of model rank correlations based on the gold and predicted \gls{nlg} performances.
Additionally, we visualize the predicted performance curves in Figure~\ref{fig:pred}.

As can be seen on Table~\ref{tab:pred}, 
all three reformulation versions perform on par with each other, achieving 6.4\% prediction error at most.
As expected, \gls{mc} and \gls{mc}$_{rnd}$ perform on par, the former being
slightly better, highlighting the advantage of smart distractors.
When looking at the average rank correlations of our 8 models we found strong
Spearman correlation values on all tasks, except GSM8K which still show moderate
correlation.
We hypothesize that this is due to the difficult nature, and the low results
(under 0.2\% on the majority of the training course; Figure~\ref{fig:pred}), of
the task, thus model ranking is more prone to noise than in case of other tasks.
Overall, predicting \gls{nlg} performance using only 3 training datapoints
proves
efficient in following generative model performance more closely,
while requiring only 
few expensive direct \gls{nlg} benchmark
evaluation, striking a balance between precise \gls{nlg} performance monitoring and compute efficiency.

\section{Conclusion}
We performed an empirical study on the relation between \gls{nlg} and \gls{nlu} benchmarks, as well as the possibility to automatically reformulate \gls{nlg} benchmarks to \gls{nlu}.
Calculating the correlation, we demonstrated that all four benchmarks we used herein had a high correlation between the \gls{nlg} and \gls{nlu} variants.
Interestingly, the correlation between these variants existed in all \gls{mc} versions that we tested\textemdash both random and smart distractors, and related off-the-shelf \gls{nlu} benchmarks\textemdash and was similarly high.
Although \gls{ll} is slightly less efficient than the \gls{mc} variants, it is still a valid option that does not need distractor option generation (even though the efforts needed are minimal).
Furthermore, we were able to show that runtime could be reduced significantly (2x–176x averaged across model sizes).
Hence, we conclude that \gls{nlu} can be used to estimate \gls{nlg} model performance to save compute, although we advice against neglecting \gls{nlg} benchmark evaluation altogether.
Furthermore, we also tested whether using only a few \gls{nlg} evaluations together with \gls{nlu} formulations is beneficial in \gls{nlg} performance prediction.
We found that using only 3 training datapoints, predicting NLG performance proves efficient for closely tracking generative model quality, reducing the need for frequent costly benchmark evaluations.

\section{Limitations}
We have shown that generative capabilities of small and medium size models can be accessed using reformulation to \gls{nlu} in various domains.
However, it is possible that bigger models (above 7B parameters) behave slightly differently, although we expect these results to hold for them as well. 
Furthermore, we did not include safety-related domains e.g., ethical evaluation benchmarks, as we consider the models used herein too small to meaningfully handle such complex tasks—capabilities more likely present in larger and/or instruction tuned models.
In such cases, the potential for task reformulation would also need to be explored.

\section*{Acknowledgments}
This work has been funded by the Free State of Bavaria in the DSgenAI project (Grant Nr.: RMF-SG20-3410-2-18-4).
The authors gratefully acknowledge the scientific support and HPC resources provided by the Erlangen National High Performance Computing Center (NHR@FAU) of the Friedrich-Alexander-Universität Erlangen-Nürnberg (FAU) under the NHR project \emph{ELMOD: Efficient language models for on-device deployment} (Grant Nr.: b239dc). NHR funding is provided by federal and Bavarian state authorities. NHR@FAU hardware is partially funded by the German Research Foundation (DFG) – 440719683.
We would like to thank our anonymous reviewers and colleagues for the useful feedback, including: Christian Kroos, Joel Schlotthauer, Lucas Druart, Luzian Hahn and Rishiraj Saha Roy.

\bibliography{custom}

\appendix

\section{Prompt for Smart Distractor Creation}
\label{sec:app:distractor_prompt}

\footnotesize
\begin{quote}
\textbf{Prompt:} \\
Please use the given question \texttt{\{question\}} and create 4 answers, the first one being \texttt{\{correct\_answer\}}, the correct answer, and the other three being incorrect answers. Use \texttt{JSONL} to respond.
\end{quote}

\textbf{Examples for TriviaQA:}
\begin{itemize}
  \item \texttt{question="Which American-born Sinclair won the Nobel Prize for Literature in 1930?",} \\
  \texttt{answers=["Sinclair Lewis", "Upton Sinclair", "Sinclair Ferguson", "Sinclair Smith"]}
  
  \item \texttt{question="Where in England was Dame Judi Dench born?",} \\
  \texttt{answers=["York", "London", "Manchester", "Oxford"]}

  \item \texttt{question="When did the founder of Jehovah's Witnesses say the world would end?",} \\
  \texttt{answers=["1914", "2012", "1844", "1975"]}

  \item \texttt{question="1998 was the Chinese year of which creature?",} \\
  \texttt{answers=["tiger", "rabbit", "dragon", "giraffe"]}

  \item \texttt{question="The first credit cards were for use in what type of establishments?",} \\
  \texttt{answers=["restaurants", "cinemas", "gas stations", "hotels"]}
\end{itemize}

\textbf{Examples for GSM8K:}
\begin{itemize}
  \item \texttt{question="Natalia sold clips to 48 of her friends in April, and then she sold half as many clips in May. How many clips did Natalia sell altogether in April and May?",} \\
  \texttt{answers=[} \\
  \texttt{"Natalia sold 48/2 = \textless\textless 48/2=24\textgreater\textgreater 24 clips in May. Natalia sold 48+24 = \textless\textless 48+24=72\textgreater\textgreater 72 clips altogether in April and May. \#\#\#\# 72",} \\
  \texttt{"Natalia sold 48/2 = \textless\textless 48/2=24\textgreater\textgreater 24 clips in May. Natalia sold 48 + 20 = \textless\textless 48+20=68\textgreater\textgreater 68 clips altogether in April and May. \#\#\#\# 68",} \\
  \texttt{"Natalia sold 48/2 = \textless\textless 48/2=24\textgreater\textgreater 24 clips in May. Natalia sold 48 + 22 = \textless\textless 48+22=70\textgreater\textgreater 70 clips altogether in April and May. \#\#\#\# 70",} \\
  \texttt{"Natalia sold 48 × 2 = \textless\textless 48*2=96\textgreater\textgreater 96 clips in May. Natalia sold 48 + 96 = \textless\textless 48+96=144\textgreater\textgreater 144 clips altogether in April and May. \#\#\#\# 96"]}

  \item \texttt{question="Weng earns \$12 an hour for babysitting. Yesterday, she just did 50 minutes of babysitting. How much did she earn?",} \\
\texttt{answers=[} \\
  \texttt{"Weng earns 12/60 = \textless\textless 12/60=0.2 \textgreater\textgreater 0.2 per minute. Working 50 minutes, she earned 0.2 x 50 = \textless\textless 0.2*50=10\textgreater\textgreater 10. \#\#\#\# 10",} \\
    \texttt{"Weng earns 12/60 = \textless\textless12/60=0.2 \textgreater\textgreater 0.2 per minute. Working 50 minutes, she earned 0.2 × 60 = \textless\textless0.2*60=12 \textgreater\textgreater 12. \#\#\#\# 12",} \\
    \texttt{"Weng earns 12/60 = \textless\textless12/60=0.2 \textgreater\textgreater 0.2 per minute. Working 50 minutes, she earned 0.2 × 40 = \textless\textless0.2*40=8>>8. \#\#\#\# 8,} \\
    \texttt{"Weng earns 12/60 = \textless\textless12/60=0.2 \textgreater\textgreater 0.2 per minute. Working 50 minutes, she earned 0.2 × 45 = \textless\textless0.2*45=9 \textgreater\textgreater 9. \#\#\#\# 9"]} \\

\item \texttt{question="Betty is saving money for a new wallet which costs \$100. Betty has only half of the money she needs. Her parents decided to give her \$15 for that purpose, and her grandparents twice as much as her parents. How much more money does Betty need to buy the wallet?",} \\
\texttt{answers=[} \\
\texttt{"In the beginning, Betty has only 100 / 2 = \$\textless\textless 100/2=50 \textgreater\textgreater 50. Betty's grandparents gave her 15 * 2 = \$\textless\textless 15*2=30 \textgreater\textgreater 30. This means, Betty needs 100 - 50 - 30 - 15 = \$\textless\textless 100-50-30-15=5 \textgreater\textgreater 5 more. \#\#\#\# 5",} \\
\texttt{"In the beginning, Betty has only 100 / 2 = \$\textless\textless 100/2=50 \textgreater\textgreater 50. Betty's grandparents gave her 15 * 2 = \$\textless\textless 15*2=30 \textgreater\textgreater 30. This means, Betty needs 100 - 50 - 30 = \$\textless\textless 100-50-30=20 \textgreater\textgreater 20 more. \#\#\#\# 20",} \\
\texttt{"In the beginning, Betty has only 100 / 2 = \$\textless\textless 100/2=50 \textgreater\textgreater 50. Betty's grandparents gave her 15 * 2 = \$\textless\textless 15*2=30 \textgreater\textgreater 30. This means, Betty needs 100 - 50 - 15 = \$\textless\textless 100-50-15=35 \textgreater\textgreater 35 more. \#\#\#\# 35",} \\
\texttt{"In the beginning, Betty has only 100 / 2 = \$\textless\textless 100/2=50 \textgreater\textgreater 50. Betty's grandparents gave her 15 * 2 = \$\textless\textless 15*2=30 \textgreater\textgreater 30. This means, Betty needs 100 - 30 - 15 = \$\textless\textless 100-30-15=55 \textgreater\textgreater 55 more. \#\#\#\# 55"]} \\

\item \texttt{question="James writes a 3-page letter to 2 different friends twice a week. How many pages does he write a year?",} \\
\texttt{answers=[} \\
\texttt{"He writes each friend 3*2= \textless\textless 3*2=6 \textgreater\textgreater 6 pages a week. So he writes 6*2= \textless\textless 6*2=12 \textgreater\textgreater 12 pages every week. That means he writes 12*52= \textless\textless 12*52=624 \textgreater\textgreater 624 pages a year. \#\#\#\# 624",} \\
\texttt{"He writes each friend 3×2 = \textless\textless 32=6 \textgreater\textgreater 6 pages a week.. So he writes 6×2 = \textless\textless 62=12 \textgreater\textgreater 12 pages every week. That means he writes 12×50 = \textless\textless 12*50=600 \textgreater\textgreater 600 pages a year. \#\#\#\# 600",} \\
\texttt{"He writes each friend 3×2 = \textless\textless 32=6 \textgreater\textgreater 6 pages a week. So he writes 6×1 = \textless\textless 61=6 \textgreater\textgreater 6 pages every week. That means he writes 6×52 = \textless\textless 6*52=312 \textgreater\textgreater 312 pages a year. \#\#\#\# 312",} \\
\texttt{"He writes each friend 3×2 = \textless\textless 32=6 \textgreater\textgreater 6 pages a week. So he writes 6×2 = \textless\textless 62=12 \textgreater\textgreater 12 pages every week. That means he writes 12×12 = \textless\textless 12*12=144 \textgreater\textgreater 144 pages a year. \#\#\#\# 144"]} \\

\end{itemize}

\normalsize

\begin{table*}[t]
\centering
\begin{tabular}{ll|rrr}
	NLG                                                        & NLU           & $P_{macro}$ & $P_{micro}$ & Spearman \\
	\hline
	\hline
  \multirow{5}{*}{\rotatebox{90}{GSM8K}}                     & \gls{mc}           & \tablevaluewithsdsix{0.75}{0.12}{0.52}{0.00}{0.76}{0.03} \\
                                                             & \gls{mc}$_{rnd}$   & \tablevaluewithsdsix{0.76}{0.11}{0.57}{0.00}{0.76}{0.03} \\
                                                             & \gls{ll}           & \tablevaluewithsdsix{0.79}{0.09}{0.56}{0.00}{0.81}{0.01} \\
  \cline{2-5}                                                & \gls{mc}$_{ao}$    & \tablevaluewithsdsix{0.38}{0.26}{0.90}{0.00}{0.88}{0.00} \\
                                                             & \gls{ll}$_{ao}$    & \tablevaluewithsdsix{0.22}{0.38}{-0.04}{0.80}{-0.07}{0.87} \\
	\hline
  \multirow{4}{*}{\rotatebox{90}{Trivia}}                    & \gls{mc}           & \tablevaluewithsdsix{0.90}{0.03}{0.94}{0.00}{0.86}{0.01} \\
                                                             & \gls{mc}$_{rnd}$   & \tablevaluewithsdsix{0.91}{0.02}{0.88}{0.00}{0.98}{0.00} \\
                                                             & \gls{ll}           & \tablevaluewithsdsix{0.90}{0.03}{0.69}{0.00}{0.81}{0.01} \\
                                                             & MMLU               & \tablevaluewithsdsix{0.89}{0.04}{0.94}{0.00}{0.95}{0.00} \\
	\hline
  \multirow{4}{*}{\rotatebox{90}{SQuAD}}                     & \gls{mc}$_{rnd}$   & \tablevaluewithsdsix{0.90}{0.03}{0.88}{0.00}{0.93}{0.00} \\
                                                             & \gls{mc}$_{rnd^*}$ & \tablevaluewithsdsix{0.90}{0.03}{0.84}{0.00}{0.93}{0.00} \\
                                                             & \gls{ll}           & \tablevaluewithsdsix{0.65}{0.15}{0.85}{0.00}{0.69}{0.06} \\
                                                             & BoolQ              & \tablevaluewithsdsix{0.78}{0.11}{0.24}{0.09}{0.45}{0.26} \\
	\hline
  \multirow{3}{*}{\rotatebox{90}{\parbox{1.1cm}{Human\\Eval}}} & \gls{mc}           & \tablevaluewithsdsix{0.83}{0.09}{0.79}{0.00}{0.81}{0.02} \\
	                                                           & \gls{mc}$_{rnd}$   & \tablevaluewithsdsix{0.85}{0.07}{0.75}{0.00}{0.81}{0.02} \\
	                                                           & \gls{ll}           & \tablevaluewithsdsix{0.86}{0.07}{0.73}{0.00}{0.79}{0.03} \\
	\hline
  \multirow{3}{*}{\rotatebox{90}{cpp}}                       & \gls{mc}           & \tablevaluewithsdsix{0.87}{0.06}{0.86}{0.00}{0.74}{0.04} \\
                                                             & \gls{mc}$_{rnd}$   & \tablevaluewithsdsix{0.92}{0.03}{0.83}{0.00}{0.90}{0.00} \\
                                                             & \gls{ll}           & \tablevaluewithsdsix{0.92}{0.03}{0.77}{0.00}{0.81}{0.01} \\
	\hline
  \multirow{3}{*}{\rotatebox{90}{go}}                        & \gls{mc}           & \tablevaluewithsdsix{0.67}{0.22}{0.71}{0.00}{0.76}{0.03} \\
                                                             & \gls{mc}$_{rnd}$   & \tablevaluewithsdsix{0.68}{0.21}{0.74}{0.00}{0.81}{0.01} \\
                                                             & \gls{ll}           & \tablevaluewithsdsix{0.66}{0.24}{0.72}{0.00}{0.88}{0.00} \\
	\hline
  \multirow{3}{*}{\rotatebox{90}{java}}                      & \gls{mc}           & \tablevaluewithsdsix{0.85}{0.10}{0.86}{0.00}{0.98}{0.00} \\
                                                             & \gls{mc}$_{rnd}$   & \tablevaluewithsdsix{0.86}{0.08}{0.80}{0.00}{0.83}{0.01} \\
                                                             & \gls{ll}           & \tablevaluewithsdsix{0.90}{0.05}{0.64}{0.00}{0.67}{0.07} \\
	\hline
  \multirow{3}{*}{\rotatebox{90}{js}}                        & \gls{mc}           & \tablevaluewithsdsix{0.92}{0.02}{0.80}{0.00}{0.83}{0.01} \\
                                                             & \gls{mc}$_{rnd}$   & \tablevaluewithsdsix{0.86}{0.06}{0.79}{0.00}{0.88}{0.00} \\
                                                             & \gls{ll}           & \tablevaluewithsdsix{0.90}{0.04}{0.75}{0.00}{0.71}{0.05} \\
	\hline
  \multirow{3}{*}{\rotatebox{90}{python}}                    & \gls{mc}           & \tablevaluewithsdsix{0.92}{0.02}{0.81}{0.00}{0.74}{0.04} \\
                                                             & \gls{mc}$_{rnd}$   & \tablevaluewithsdsix{0.91}{0.02}{0.81}{0.00}{0.79}{0.02} \\
                                                             & \gls{ll}           & \tablevaluewithsdsix{0.93}{0.01}{0.81}{0.00}{0.81}{0.01} \\
	\hline
  \multirow{3}{*}{\rotatebox{90}{rust}}                      & \gls{mc}           & \tablevaluewithsdsix{0.77}{0.14}{0.67}{0.00}{0.79}{0.02} \\
                                                             & \gls{mc}$_{rnd}$   & \tablevaluewithsdsix{0.87}{0.04}{0.52}{0.00}{0.62}{0.10} \\
                                                             & \gls{ll}           & \tablevaluewithsdsix{0.87}{0.05}{0.70}{0.00}{0.83}{0.01} \\
\end{tabular}\caption{
Correlation statistics of the \gls{nlg} tasks and their various reformulated formats.
We present p-values in parentheses.
The \gls{nlg} task \emph{Code} represents HumanEval results averaged over the 6 coding languages (cpp, go, java, js, python and rust).
\gls{mc} stands for the multiple-choice formulation with smart distractors, \gls{mc}$_{rnd}$ uses random distractors, while \gls{ll} is the log-likelihood formulation.
For GSM8K we tested setups where only the final answer has to be scored by the model (*$_{ao}$) as proposed by \cite{gsm8k-mc}.
In case of TriviaQA and SQuAD, we tested  cross-benchmark pairings: MMLU and BoolQ respectively.
Additionally, in \gls{mc}$_{rnd^*}$ we used more than 4 answer options for SQuAD.
}
\label{tab:all_results}
\end{table*}

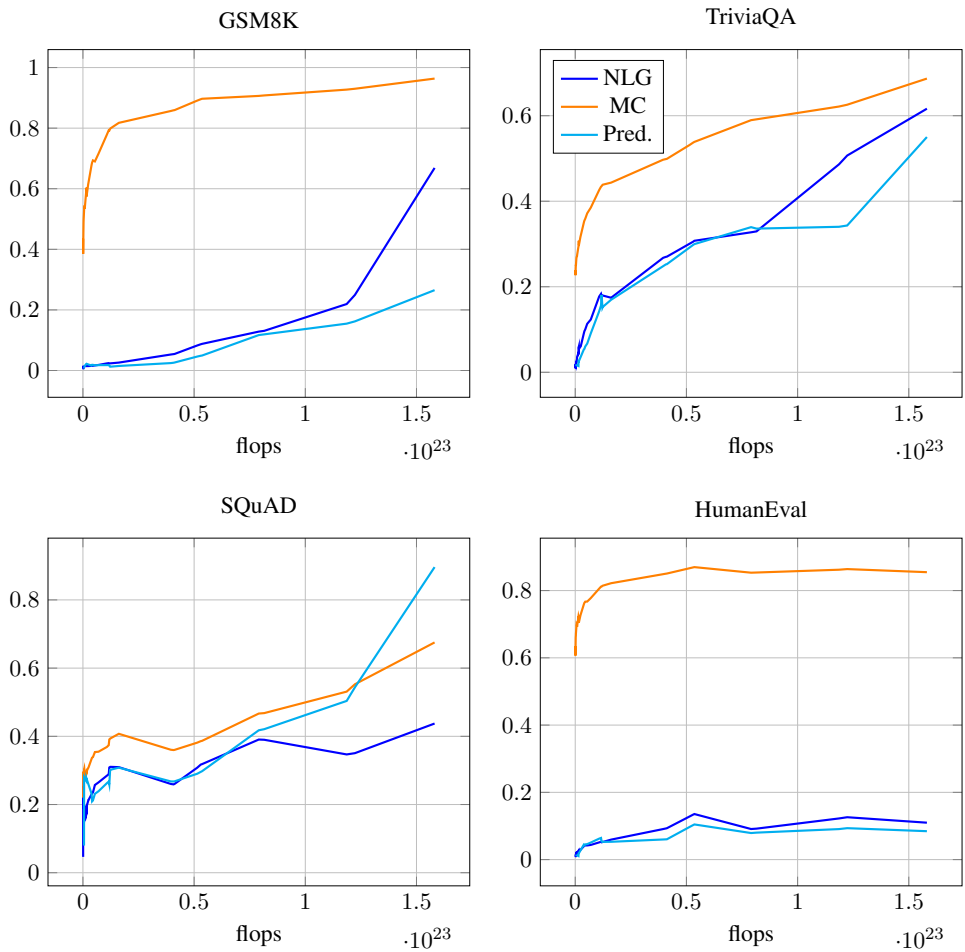
\begin{figure*}[t]
    \centering
    \resizebox{.8\textwidth}{!}{    \begin{tikzpicture}
            \begin{axis}[
            xlabel={flops},
                        title={GSM8K},
            grid=major,
            legend style={at={(0.5,0.5)},anchor=center},              unbounded coords=jump,
        ]
        \addplot[blue, line width=1pt,] table [x=flops, y=gsm8k, col sep=comma] {flop_task.csv};
                \addplot[orange, line width=1pt,] table [x=flops, y=gsm8k_mc_cloze_cot, col sep=comma] {flop_task.csv};
                \addplot[orange, dashed, line width=1pt] table [x=flops, y=gsm8k_mc_cloze_cot->gsm8k, col sep=comma] {flop_task.csv};
                \addplot[red, line width=1pt,] table [x=flops, y=gsm8k_ll_correct_prob_norm, col sep=comma] {flop_task.csv};
        \addplot[red, dashed, line width=1pt] table [x=flops, y=gsm8k_ll_correct_prob_norm->gsm8k, col sep=comma] {flop_task.csv};
        \end{axis}
        
            \begin{axis}[
            xlabel={flops},
                        title={TriviaQA},
            grid=major,
                        legend style={at={(0.8,0.45)},anchor=center},              at={(8cm,0)},
            unbounded coords=jump,
        ]
        \addplot[blue, line width=1pt,] table [x=flops, y=triviaqa, col sep=comma] {flop_task.csv};
        \addlegendentry{NLG};
        \addplot[orange, line width=1pt,] table [x=flops, y=triviaqa_mc_cloze, col sep=comma] {flop_task.csv};
        \addlegendentry{\gls{mc}};
        \addplot[orange, dashed, line width=1pt] table [x=flops, y=triviaqa_mc_cloze->triviaqa, col sep=comma] {flop_task.csv};
        \addlegendentry{Pred. MC};
        \addplot[red, line width=1pt] table [x=flops, y=triviaqa_ll_correct_prob_norm, col sep=comma] {flop_task.csv};
        \addlegendentry{LL};
        \addplot[red, dashed, line width=1pt] table [x=flops, y=triviaqa_ll_correct_prob_norm->triviaqa, col sep=comma] {flop_task.csv};
        \addlegendentry{Pred. LL};
        \end{axis}

            \begin{axis}[
            xlabel={flops},
                        title={SQuAD},
            grid=major,
            legend pos=north west,              at={(0cm,-8cm)},
            unbounded coords=jump,
        ]
        \addplot[blue, line width=1pt,] table [x=flops, y=squadv2_norm, col sep=comma] {flop_task.csv};
                \addplot[orange, line width=1pt] table [x=flops, y=squadv2_mc_4, col sep=comma] {flop_task.csv};
                \addplot[orange, dashed,, line width=1pt] table [x=flops, y=squadv2_mc_4->squadv2_norm, col sep=comma] {flop_task.csv};
                \addplot[red, line width=1pt] table [x=flops, y=squadv2_ll_correct_prob_norm, col sep=comma] {flop_task.csv};
        \addplot[red, dashed,, line width=1pt] table [x=flops, y=squadv2_ll_correct_prob_norm->squadv2_norm, col sep=comma] {flop_task.csv};
        \end{axis}

            \begin{axis}[
            xlabel={flops},
                        title={HumanEval},
            grid=major,
            legend style={at={(0.8,0.5)},anchor=center},              at={(8cm,-8cm)},
            unbounded coords=jump,
        ]
        \addplot[blue, line width=1pt,] table [x=flops, y=humanevalsynthesize_avg, col sep=comma] {flop_task.csv};
                \addplot[orange, line width=1pt,] table [x=flops, y=humanevalpack_cloze_debug_avg, col sep=comma] {flop_task.csv};
                \addplot[orange, dashed, line width=1pt] table [x=flops, y=humanevalpack_cloze_debug_avg->humanevalsynthesize_avg, col sep=comma] {flop_task.csv};
                \addplot[red, line width=1pt] table [x=flops, y=humanevalpack_ll_correct_prob_norm_avg, col sep=comma] {flop_task.csv};
        \addplot[red, dashed, line width=1pt] table [x=flops, y=humanevalpack_ll_correct_prob_norm_avg->humanevalsynthesize_avg, col sep=comma] {flop_task.csv};
        \end{axis}
    \end{tikzpicture}
    }
    \caption{
    Predicted (Pred.) generative performance using \gls{mc} or \gls{ll} performance.
    Each predicted point is based on 3 previous \gls{nlu} and NLG reference points using linear regression.
    }
    \label{fig:pred}
\end{figure*}

\begin{table*}[t]
\centering
\resizebox{\textwidth}{!}{{\renewcommand{\arraystretch}{2.5}
\begin{tabular}{r|rrl}
model                          & \#params. & \#tokens & checkpoints                                                                                                                                                                                                                                                                                          \\ \hline
EleutherAI/pythia-1b-deduped   & 1B        & 300B     & step10000, step20000, step30000, step50000, step143000                                                                                                                                                                                                                                               \\
EleutherAI/pythia-2.8b-deduped & 2.8B      & 300B     & step10000, step20000, step30000, step50000, step143000                                                                                                                                                                                                                                               \\
EleutherAI/pythia-6.9b-deduped & 6.9B      & 300B     & step10000, step20000, step30000, step50000, step143000                                                                                                                                                                                                                                               \\
allenai/OLMo-1B-0724-hf        & 1B        & 3.05T    & \makecell[l]{step10000-tokens20B, step20000-tokens41B, step29000-tokens60B, \\ step50000-tokens104B, step150000-tokens314B, step500000-tokens1048B, \\ step621000-tokens1301B, step1000000-tokens2096B, step1454000-tokens3048B}                                                                                       \\
allenai/OLMo-7B-0724-hf        & 7B        & 2.75T    & \makecell[l]{step10000-tokens41B, step14500-tokens60B, step25500-tokens106B, \\ step75000-tokens314B, step250000-tokens1048B, step310000-tokens1300B, \\ step500000-tokens2097B, step650650-tokens2729B}                                                                                                                 \\
LLM360/Amber                   & 7B        & 1.25T    & ckpt\_005, ckpt\_010, ckpt\_016, ckpt\_027, ckpt\_082, ckpt\_275, ckpt\_358                                                                                                                                                                                                                          \\
LLM360/Crystal                 & 7B        & 1.4T     & \makecell[l]{CrystalCoder\_phase1\_checkpoint\_006000, CrystalCoder\_phase1\_checkpoint\_009000, \\ CrystalCoder\_phase1\_checkpoint\_012000, CrystalCoder\_phase1\_checkpoint\_021000, \\ CrystalCoder\_phase1\_checkpoint\_067500, CrystalCoder\_phase2\_checkpoint\_015000, \\ CrystalCoder\_phase3\_checkpoint\_027728} \\
allenai/OLMo-2-1124-7B         & 7B        & 4T       & \makecell[l]{stage1-step10000-tokens42B, stage1-step14000-tokens59B, stage1-step25000-tokens105B, \\ stage1-step75000-tokens315B, stage1-step250000-tokens1049B, stage1-step310000-tokens1301B, \\ stage1-step500000-tokens2098B, stage1-step720000-tokens3020B, stage2-ingredient1-step11931-tokens50B }               
\end{tabular}}
}
\caption{
Details of the used models.
The model and checkpoint names are references to the content on the Huggingface Hub, while the number of parameters and training tokens are based on the respective model publications.
}
\label{tab:model_ckpts}
\end{table*}

\clearpage
\onecolumn
\begin{longtable}{ll|p{4cm}|p{4cm}|p{3cm}}
\textbf{} & \textbf{benchm.} & \textbf{(correct) answer} & \textbf{distractors} & \textbf{specifics} \\ \hline \hline
\endfirsthead
\multicolumn{5}{c}{{\tablename\ \thetable{} -- continued from previous page}} \\ \hline
\textbf{} & \textbf{benchm.} & \textbf{answer} & \textbf{distractors} & \textbf{specifics} \\ \hline
\endhead
\hline \multicolumn{5}{r}{{Continued on next page}} \\
\endfoot
\endlastfoot

\multirow{10}{*}{\raisebox{-0.5\totalheight}{\rotatebox[origin=c]{90}{GSM8K}}} & question & \multicolumn{3}{p{12cm}}{Natalia sold clips to 48 of her friends in April, and then she sold half as many clips in May. How many clips did Natalia sell altogether in April and May?} \\
\cline{2-5}
 & original & \multirow{11}{=}{Natalia sold 48/2 = \textless{}\textless{}48/2=24\textgreater{}\textgreater{}24 clips in May. Natalia sold 48+24 = \textless{}\textless{}48+24=72\textgreater{}\textgreater{}72 clips altogether in April and May. \#\#\#\# 72} & & \\ \cline{4-5}
 & \gls{mc} & & {[}Natalia sold 48 × 2 =   \textless{}\textless{}48*2=96\textgreater{}\textgreater{}96 clips in May. Natalia sold 48 + 96 =   \textless{}\textless{}48+96=144\textgreater{}\textgreater{}144 clips altogether in April and May. \#\#\#\# 144,   Natalia sold {[}...{]}{]} & \\  \cline{4-5}
 & \gls{mc}$_{rnd}$ & & {[}Weng earns 12/60 = <<12/60=0.2>>..., He writes each friend {[}…{]}{]} & \\ \cline{2-5}
 & GSM-MC & 72 & {[}64, 61, 89{]} & \\ 
\hline

\multirow{5}{*}{\raisebox{-0.5\totalheight}{\rotatebox{90}{Trivia}}} & question & \multicolumn{3}{p{12cm}}{Which was the first European country to abolish capital punishment?} \\
\cline{2-5}
 & original & \multirow{10}{=}{Norway} & & aliases: Norvège, Mainland Norway, Norwegian state, {[}…{]} normalized\_aliases: norwegen, kongeriket norge, norway, {[}…{]} \\ \cline{4-5}
 & \gls{mc} &  & {[}Germany, Italy, Poland{]} & \\ \cline{4-5}
 & \gls{mc}$_{rnd}$ &  & {[}Chicago Bears, Ballet, 6{]} & \\
\hline

\multirow{15}{*}{\raisebox{-0.5\totalheight}{\rotatebox{90}{SQuAD}}} & question & \multicolumn{3}{p{12cm}}{Who is the main character in "Childe Harold's: Canto I?"} \\
\cline{2-5}
 & original & no answer in this context & & context: {[}…{]} Studying and analyzing literature becomes very important in terms of learning about our history. {[}…{]} Lord Byron talks about the Spanish and the French in ‘‘Childe Harold’s Pilgrimage: Canto I’’ {[}…{]} \\  \cline{2-5}
 & question & \multicolumn{3}{p{12cm}}{We can learn what by carefully examining our literature?} \\ 
\cline{2-5}
 & original & \multirow{15}{=}{our history} & & context: {[}…{]} Studying and analyzing literature becomes very important in terms of learning about our history. {[}…{]} Lord Byron talks about the Spanish and the French in ‘‘Childe Harold’s Pilgrimage: Canto I’’ {[}…{]} \\ \cline{4-5}
 & \gls{mc}$_{rnd}$ & & {[}Lord Byron, written records, corpse{]} & answers taken from same context \\ \hline

\multirow{9}{*}{\raisebox{-0.5\totalheight}{\rotatebox{90}{HumanEval}}} & question & \multicolumn{3}{p{9cm}}{[...] Given a string, find out how many distinct characters (regardless of case) does it consist of [...]} \\
\cline{2-5}
 & original & \multirow{7}{=}{return len(set(string.lower()))} & & buggy solution: return set(len(string.lower())) \\ \cline{4-5}
 & \gls{mc} &  & return set(len(string.lower())) &  \\ \cline{4-5}
  & \gls{mc}$_{rnd}$ &  & [return ' '.join([str(x) for x in range(n + 1)]), [...]] &  \\
\hline \hline

 &\multicolumn{4}{l}{MMLU} \\ \cline{2-5}
 & question & \multicolumn{3}{p{9cm}}{Find the degree for the given field extension Q(sqrt(2), sqrt(3), sqrt(18)) over Q.} \\
\cline{2-5}
 &  & 4 & {[}0,2,6{]} & \\ \hline
 &\multicolumn{4}{l}{BoolQ} \\ \cline{2-5}
 & question & \multicolumn{3}{p{9cm}}{do iran and afghanistan speak the same language} \\
\hline
 &  & TRUE & FALSE & \\
\caption{
Example items from the used benchmarks and our reformulations.
GSM-MC references \cite{gsm8k-mc}. As MMLU and BoolQ were used as \gls{mc} pairings of other benchmarks, they only have the original version.
}
\label{tab:example_all_benchmarks}
\end{longtable}
\clearpage
\twocolumn

\end{document}